\documentclass[sigconf]{acmart}
\AtBeginDocument{%
  }

\copyrightyear{2026}
\acmYear{2026}
\setcopyright{cc}
\setcctype{by}
\acmConference[WSDM '26]{Proceedings of the Nineteenth ACM International Conference on Web Search and Data Mining}{February 22--26, 2026}{Boise, ID, USA}
\acmBooktitle{Proceedings of the Nineteenth ACM International Conference on Web Search and Data Mining (WSDM '26), February 22--26, 2026, Boise, ID, USA}
\acmPrice{}
\acmDOI{10.1145/3773966.3777999}
\acmISBN{979-8-4007-2292-9/2026/02}

\graphicspath{{images/}}
\usepackage{subcaption}
\usepackage{adjustbox}

\usepackage{enumitem}
\usepackage{balance}

\newcommand{\myparagraph}[1]{\vspace{3pt}\noindent{\bf #1}}

\def\CVAE{{\text{UBCG}}}
\def\CVAEfull{{Universal Bimodal Conditional Generator}}

\begin{document}

\title{Prompt Tuning without Labeled Samples for Zero-Shot Node Classification in Text-Attributed Graphs}

\author{Sethupathy Parameswaran}
\authornote{Part of this work was carried out while working as a Research Engineer at Singapore Management University}
\affiliation{%
  \institution{Indian Institute of Science}
  \city{Bangalore}
  \state{Karnataka}
  \country{India}
}
\email{sethupathyp@iisc.ac.in}

\author{Suresh Sundaram}
\affiliation{%
  \institution{Indian Institute of Science}
  \city{Bangalore}
  \state{Karnataka}
  \country{India}}
\email{vssuresh@iisc.ac.in}

\author{Yuan Fang}
\affiliation{%
  \institution{Singapore Management University}
  \city{Singapore}
  \country{Singapore}
}
\email{yfang@smu.edu.sg}


\begin{abstract}
Node classification is a fundamental problem in information retrieval with many real-world applications, such as community detection in social networks, grouping articles published online and product categorization in e-commerce. Zero-shot node classification in text-attributed graphs (TAGs) presents a significant challenge, particularly due to the absence of labeled data. In this paper, we propose a novel Zero-shot Prompt Tuning (ZPT) framework to address this problem by leveraging a Universal Bimodal Conditional Generator (UBCG). Our approach begins with pre-training a graph-language model to capture both the graph structure and the associated textual descriptions of each node. Following this, a conditional generative model is trained to learn the joint distribution of nodes in both graph and text modalities, enabling the generation of synthetic samples for each class based solely on the class name. These synthetic node and text embeddings are subsequently used to perform continuous prompt tuning, facilitating effective node classification in a zero-shot setting. Furthermore, we conduct extensive experiments on multiple benchmark datasets, demonstrating that our framework performs better than existing state-of-the-art baselines. We also provide ablation studies to validate the contribution of the bimodal generator. The code is provided at: \url{https://github.com/Sethup123/ZPT}.
\end{abstract}




\maketitle

\section{Introduction}\label{sec:intro}

\begin{figure}[t]
	\centering
	\includegraphics[width=0.99\linewidth]{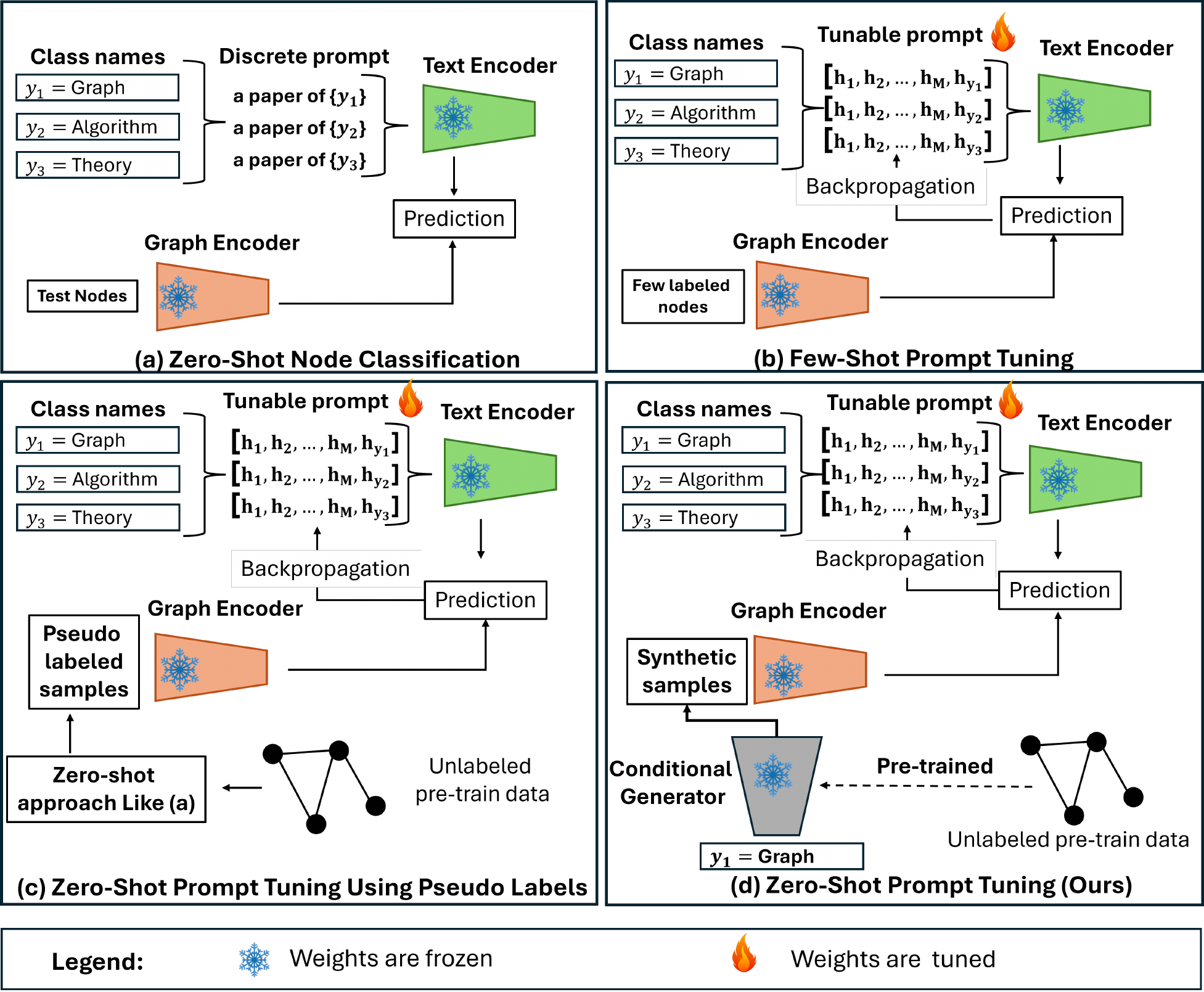}
	\vspace{-2mm}
    \caption{Illustration of different prompt learning techniques. The frozen symbol indicates the parameters are frozen; the fire symbol indicates that the parameters are tuned.}
	\label{fig:intro}
\end{figure}

Graphs underpin many essential applications on the Web, such as user relationships in social networks \cite{zhang2022robust, zhou2023hierarchical, fan2019graph}, user-item interactions in e-commerce platforms \cite{web_scale_recommender_systemsying2018graph, qu2023semi}, and citations in hyperlinked articles or scientific papers \cite{kanakia2019scalable, wang2022disencite}.
In these graphs, node classification \cite{wu2022nodeformer, wang2020nodeaug, wei2023clnode}---assigning meaningful labels to nodes such as users or articles---enables more intelligent and efficient systems.
For example, in social networks, classifying users based on their interests or behavior can enhance content recommendation, while in citation graphs, assigning topics to articles aids in trend analysis and knowledge discovery. 
%
Conventional supervised or semi-supervised models for node classification often suffer from two drawbacks. First, they require a large amount of labeled nodes for training, and acquiring human-annotated labels can be difficult or expensive. Second, such a supervised paradigm cannot cope with novel classes that are unseen during training, limiting their generalization capabilities. 
Hence, this motivates the need for approaches that can perform node classification for arbitrary classes without relying on labeled nodes. 

\myparagraph{Research problem.} In this work, we address the problem of \emph{zero-shot node classification} on graphs, where we only have access to class names without any labeled nodes. A class name is a short textual phrase describing the class or category, such as ``theory'' or ``algorithm,'' for computer science articles in a citation graph. 
More specifically, our work focuses on \emph{text-attributed graphs} (TAGs) \cite{zhaolearning, yan2023comprehensive}, a common graph structure where each node is associated with a textual description, which provides semantic signals relevant to class names. 
TAGs naturally arise in various applications. In a user-item interaction graph for e-commerce, each item has a title and a product description, while in a citation graph, each article is associated with a title and an abstract or a main text. 

Towards zero-shot node classification on TAGs, pre-training a graph-language model \cite{g2p2_wen2023augmenting, Hound_wang2024hound} has emerged as a state-of-the-art paradigm. These methods first perform self-supervised learning on unlabeled TAGs to pre-train a graph-language model. Subsequently, they design discrete prompts \cite{schick2021s, gpt3_brown2020language} for the pre-trained model to infer node class memberships in a zero-shot setup, as illustrated in Fig.~\ref{fig:intro}a.
However, discrete prompts are handcrafted and lack a principled approach to ensure optimal performance. On the other hand, continuous prompts \cite{li2021prefix, lester2021power} are learnable token vectors that eliminate the need for manual engineering. They are optimized for each classification task through \emph{prompt tuning}, requiring at least one labeled example per class, as illustrated in Fig.~\ref{fig:intro}b. That means continuous prompt tuning work only for one- or few-shot classification and not for zero-shot classification.


A na\"ive workaround to enable prompt tuning for a zero-shot classification task is to use existing zero-shot methods, such as discrete prompt-based approaches, to assign pseudo class labels to unlabeled nodes in pre-training graphs. These pseudo labels can then be used for prompt tuning, simulating the availability of labeled nodes, as shown in Fig.~\ref{fig:intro}c. However, errors in the pseudo labels are hard to control and may be further amplified during prompt tuning. Moreover, when a new classification task with novel classes emerges, pseudo labels must be regenerated, which makes scaling difficult.

\myparagraph{Insights.} To overcome the above issues, we address the following research question for node classification on TAGs: \emph{How can zero-shot prompt-tuning be performed without relying on any label or pseudo label}?
%
Generative models \cite{kingma2013auto, goodfellow2020generative, ho2020denoising} are known for learning the underlying data distribution and are capable of generating synthetic samples. Thus, we propose training a conditional generative model \cite{mirza2014conditional, pumarola2020c}, which can then generate class-specific synthetic nodes by conditioning on class names for a zero-shot classification task. These synthetic nodes subsequently serve as labeled samples for prompt tuning. 
However, the design of the conditional generative model is non-trivial with two key desiderata. 

First, to ensure scalability, it must be trained only once following the pre-training of the graph-language model so that it is universally applicable to all downstream classification tasks, rather than being retrained for each task. This implies that it only has access to unlabeled pre-training graphs, with neither labeled nodes nor class names, as it cannot anticipate arbitrary future classification tasks. Hence, during the training of the generative model, we cannot directly use the class name as the conditioning input to reconstruct its node sample. Instead, we propose using the textual description of each node as a proxy for its class name, since the textual descriptions of nodes belonging to the same class are likely to share a strong semantic similarity. Based on this proxy, the conditional generative model can be trained without any labeled samples; during inference, it can still leverage class names as conditioning input to generate synthetic node samples for each class in a downstream classification task, as shown in Fig.~\ref{fig:intro}d.

Second, the generative model must be bimodal, capable of generating both graph and text representations of the same node to facilitate the fusion of both modalities, as synthetic nodes with both modalities can improve the performance of prompt tuning.   
To achieve this, we train the conditional generative model in two ways: not only conditioning on a node's textual representation to generate its graph representation, but also conditioning on its graph representation to generate its textual representation. Hence, during inference, conditioned on the textual representation of a class name, we first generate a synthetic graph representation of a node for the corresponding class. Then, conditioned on the synthetic graph representation, we generate the corresponding synthetic textual representation. 
Hence, synthetic representations of both modalities will be used to optimize a continuous prompt. 

It is worth noting that our conditional generative model operates entirely in the latent space in which the input, conditioning and output are all represented as continuous embeddings. Such latent representations are generally smoother than discrete pseudo-label assignments, which may attenuate the impact of labeling errors.

\myparagraph{Contributions.}
Based on the above insights, we propose a Zero-shot Prompt Tuning (ZPT) approach for node classification on TAGs without any labeled samples. 
Specifically, we first pre-train an off-the-shelf graph-language model for our graph to obtain both graph and textual representations of each node. 
Next, as our core contribution, a \CVAEfull\ (\CVAE) is trained to capture the underlying distribution of the nodes in both graph and textual modalities.
Subsequently, given a downstream zero-shot node classification task, where only class names are available, we use the trained \CVAE\ to produce synthetic graph and textual representations for each class.  
Finally, these multi-modal synthetic samples are used to perform continuous prompt tuning for the classification task.

In summary, our work makes the following key contributions.
 (1) We investigate a novel zero-shot prompt tuning setting for node classification on TAGs, which does not require labeled samples at any stage.
  (2)  We propose a novel conditional generative model, \CVAE, that eliminates the need for both labeled nodes and class names, allowing it to be trained only once and remain universally applicable to any unseen class that may emerge in future downstream tasks. Moreover, it can generate synthetic representations in both graph and textual modalities to further enhance prompt tuning.
   (3) Our experimental results show that the proposed method, ZPT, achieves superior performance across four benchmark datasets compared to a wide range of state-of-the-art approaches. 

\section{Related Work}

\myparagraph{Language pre-training and prompting.} The field of language pre-training and prompting has rapidly evolved, with foundational models like BERT \cite{kenton2019bert}, GPT \cite{radford2018improving}, and T5 \cite{raffel2020exploring} pioneering the use of large-scale pre-training on extensive text corpora. These models capture rich linguistic and contextual knowledge during pre-training, enabling effective transfer to various downstream tasks through fine-tuning or prompting. Early work, such as BERT's masked language model and GPT's autoregressive pre-training, demonstrated the strength of this paradigm in tasks like text classification \cite{yang2019xlnet, sun2019ernie}. Prompt-based learning, introduced in models like GPT-3 \cite{gpt3_brown2020language}, extended this idea by using prompts to frame tasks, allowing models to perform tasks with few-shot or even zero-shot learning \cite{prompt_tuning_wang2022towards, hu2022knowledgeable, zhong2021adapting}. This method minimizes the need for task-specific data and fine-tuning, making it a powerful approach for generalization across a wide range of applications.

\myparagraph{Graph pre-training and prompting.} Graph neural networks (GNNs) \cite{GCN_kipf2016semi, GAT_velivckovic2017graph, variational_graph_autoencoders_kipf2016variational} operate by propagating information through the graph structure, where features from neighboring nodes are aggregated through message-passing mechanisms to compute the embedding for each node in the graph. However, traditional supervised methods like GCN \cite{GCN_kipf2016semi}, GAT \cite{GAT_velivckovic2017graph}, and GIN \cite{GIN_xu2018powerful} require many labeled samples for training. Another branch of methods is the self-supervised methods \cite{DGI_velivckovicdeep, graphcl_you2020graph, graphmae_hou2022graphmae, GCA_zhu2021graph, graphmae2_hou2023graphmae2, mgae_wang2017mgae}, which make use of supervisory signals from the data using well defined pretext tasks without requiring labeled samples. Specifically, DGI \cite{DGI_velivckovicdeep} maximizes the mutual information between local node representations and a global graph-level summary, encouraging the model to capture meaningful structural and feature information. GraphCL \cite{graphcl_you2020graph} learns robust graph embeddings by maximizing agreement between different augmented views of the same graph while minimizing agreement between views of different graphs. In GraphMAE \cite{graphmae_hou2022graphmae}, a portion of the node features is masked, and the model is tasked with reconstructing the masked features based on the surrounding graph structure and remaining node features. Even though self-supervised approaches can learn robust embeddings, they still require labeled samples for adapting to downstream tasks such as node classification. Few-shot node classification methods \cite{graph_few_shot_yao2020graph, Graph_prototypical_few_shot_ding2020graph, Meta-gnn_zhou2019meta} address the requirement of large amount of labeled samples by learning to categorize nodes within a graph using only a few labeled samples. Furthermore, 
several studies have adopted the pre-training and prompt tuning paradigm to tackle the challenge of few-shot node classification on graphs \cite{universl_prompt_fang2024universal, liu2023graphprompt, sun2022gppt, all_in_one_sun2023all}.

\myparagraph{Zero-shot node classification.} Zero-shot node classification has gained increasing attention as a method for classifying nodes in graphs without task-specific labeled data, leveraging the knowledge encoded in pre-trained models to generalize across unseen classes. Particularly, methods like DGPN \cite{DGPN_wang2021zero}, DBiGCN \cite{yue2022dual}, and GraphCEN \cite{ju2023zero} perform zero-shot node classification by first training on a set of labeled base classes (seen) and leverage side information such as Wikipedia pages to generalize to novel unseen classes. Recently, foundation models like  OFA \cite{one_for_all_liuone} and GraphGPT \cite{graphgpt_tang2024graphgpt} can also perform zero-shot node classification. However, similar to the above discussed methods, OFA also requires labeled base classes to generalize to unseen classes. On the other hand, GraphGPT trains on a labeled source dataset such as ArXiv and generalizes to a different target dataset such as Cora in a zero-shot fashion. However, labeled samples are still required for training before the zero-shot transfer. Different from the above methods, approaches like G2P2 \cite{g2p2_wen2023augmenting} and Hound \cite{Hound_wang2024hound} make use of unlabeled nodes and the associated textual description in text-attributed graphs (TAGs) for pre-training and perform zero-shot node classification in the downstream tasks using just the class names as discrete prompts. In particular, G2P2 jointly pre-trains a graph encoder and a text encoder and aligns the two modalities using several contrastive losses. In addition to the node-text alignment, Hound utilizes additional self-supervised signals, namely, node perturbation, text matching and semantic negation. In the downstream tasks, both G2P2 and Hound leverage the class name as discrete prompts and perform zero-shot classification. However, they require at least one labeled sample for prompt tuning. In our work, we propose a novel zero-shot prompt tuning approach that does not require any labeled sample, while eliminating the need for handcrafted discrete prompts.  

Zero-shot learning has also been extensively studied in other domains, most notably in image classification. One common approach involves learning a mapping from the visual space to the semantic space of class descriptions, and subsequently classifying images based on the similarity between their visual and semantic representations \cite{skorokhodovclass, xu2020attribute, cacheux2019modeling}. Another common method leverages a generative model that synthesizes samples for unseen classes, which are then used alongside the training data for seen classes to train a classifier \cite{mishra2018generative, schonfeld2019generalized, elhoseiny2021cizsl++, shen2020invertible}. However, it is important to note that these zero-shot image classification approaches require labeled training samples from some base (i.e., seen) classes---either to learn the visual-to-semantic mapping or to train the generative model for synthetic data production. In contrast, the zero-shot node classification setting examined in this paper does not assume the availability of any labeled samples.

\section{Preliminaries}

In this section, we introduce our problem statement, and outline the technical background on graph-language models.

\subsection{Problem Statement}
\label{sec:setting}
\myparagraph{Text-attributed graphs (TAGs).} 
A TAG \cite{zhaolearning, yan2023comprehensive} can be denoted as $\mathcal{G} = \{\mathcal{V}, \mathcal{E}, \mathbf{X}, \mathcal{T} \}$, where $\mathcal{V}$ represents a set of nodes, $\mathcal{E}$ a set of edges between the nodes, $\mathbf{X}\in \mathbb{R}^{|V|\times d}$ is a feature matrix for the nodes, and $\mathcal{T}$ is a set of textual descriptions for the nodes.
 For example, in a citation graph, each node $v \in \mathcal{V}$ corresponds to an article, and each edge $e \in \mathcal{E}$ represents the citing or cited relationship between the articles. A node $v$ is associated with both a feature vector $\mathbf{x}_v$ from $\mathbf{X}$ and a textual description from $\mathcal{T}$, which could represent the title, abstract or full content of the article. 

\myparagraph{Zero-shot node classification.} 
%
In our work, we adopt a pre-training framework \cite{g2p2_wen2023augmenting, Hound_wang2024hound} for zero-shot node classification, which consists of two stages: pre-training and downstream adaptation.

First, during the pre-training stage, we leverage existing work \cite{g2p2_wen2023augmenting} to pre-train a graph-language model on a given TAG. The TAG is unlabeled, containing no information related to downstream classification tasks. The details of the graph-language model are deferred to Sect.~\ref{sub:g2p2-pretraining}.

Then, in the downstream stage, we adapt the pre-trained model to work with zero-shot node classification tasks.
In each task, consider a set of $N$ classes for the nodes on the graph. Each class is given only a class name, which is a short textual phrase (i.e., a sequence of natural language tokens), such as ``machine learning'' or ``theory'' in the context of academic citation graphs.  Note that there is no labeled node given for any class.
The goal is to predict the classes for a query set consisting of unlabeled nodes, where each node belongs to one of the $N$ classes.  This is a special case of $N$-way $K$-shot classification \cite{finn2017model}, where $K=0$.


\subsection{Graph-Language Model}
\label{sub:g2p2-pretraining}
We follow Wen and Fang~\cite{g2p2_wen2023augmenting} to pre-train a graph-language model on TAGs. 
The main objective is to jointly train a graph encoder $G_{\theta}$ and a text encoder $T_{\phi}$ in an aligned space, given the correspondence between nodes and their textual descriptions. Specifically, we model the graph encoder using a GCN \cite{GCN_kipf2016semi} and the text encoder using a Transformer \cite{vaswani2017attention}. Both encoders are pre-trained on an unlabeled TAG before proceeding to classification tasks. 

\myparagraph{Pre-training.} The node embeddings $\mathbf{V}$ given by the graph encoder and the corresponding text embeddings $\mathbf{T}$ given by the text encoder are aligned using the following three contrastive losses.

First, we align the representation of a node with that of its corresponding textual description by minimizing the following contrastive loss.
\begin{equation}
    \mathcal{L}_1 = \frac{1}{2}\left(\mathtt{CE}(\Lambda_1,\mathbf{y}) + \mathtt{CE}(\Lambda_1^\top,\mathbf{y})\right),
\end{equation}
where $\Lambda_1$ is the pairwise cosine similarity matrix between the node embeddings and the corresponding text embeddings, $\textbf{y} = [1,2,..n]^\top$ is the label vector for contrastive learning, and $\mathtt{CE}$ stands for cross-entropy loss applied to $\Lambda_1$ and $\Lambda_1^\top$ in a row-wise manner.
In particular, $\Lambda_1$ is computed from the $L_2$ normalized node embedding matrix $\tilde{\textbf{V}}$ and the text embedding matrix $\tilde{\textbf{T}}$: $\Lambda_1 =  (\tilde{\textbf{V}}\tilde{\textbf{T}}^\top)\cdot \exp(\tau)$, 
where $\tau$ is a trainable parameter used for scaling the similarity scores. 

Next, given a node, the textual descriptions associated with its neighbouring nodes, as defined by the graph structure, can be viewed as a summary of the node. The textual summary embedding of node $v$ is obtained as the mean of its neighbors' text embeddings, $\mathbf{s}_v = \mathtt{Mean}(\{ \mathbf{t}_u : u \in \mathcal{N}_v$\}), where $\mathcal{N}_v$ represents the set of neighbors of $v$. Let $\tilde{\mathbf{S}}$ denote the $L_2$-normalized summary embedding matrix for all nodes. Then, 
we can align the text embedding with the summary embedding using the following contrastive loss.
\begin{equation}
    \mathcal{L}_{2} = \frac{1}{2}\left(\mathtt{CE}(\Lambda_2,\mathbf{y}) + \mathtt{CE}(\Lambda_2^\top,\mathbf{y})\right),
\end{equation}
where $\Lambda_2= (\tilde{\textbf{S}}\tilde{\textbf{T}}^\top)\cdot\exp(\tau)$ is the text-summary similarity matrix.

Similarly, we can also align node embeddings with their corresponding summary embeddings:
\begin{equation}
    \mathcal{L}_{3} = \frac{1}{2}\left(\mathtt{CE}(\Lambda_3,\mathbf{y}) + \mathtt{CE},(\Lambda_3^\top,\mathbf{y})\right)
\end{equation}
where $\Lambda_3=(\tilde{\textbf{V}}\tilde{\textbf{S}}^\top)\cdot \exp(\tau)$ is the node-summary similarity matrix.

Hence, the overall alignment loss is $\mathcal{L}_\text{align} = \mathcal{L}_{1} + \alpha (\mathcal{L}_{2} + \mathcal{L}_{3})$, where $\alpha$ is a hyperparameter that balances node-summary and text-summary alignment losses.

\myparagraph{Classification with pre-trained model.}
In an N-way zero-shot classification task, the goal is to assign one of the $N$ classes to which the target node is most similar, without having access to any labeled samples. As shown in Fig.~\ref{fig:intro}a, given a class $y \in \{1,2,..,N\}$, we can construct a class description using just the class name or a discrete prompt such as ``a paper of \{class name\}''. The classification weights for class $y$, $\mathbf{w}_y$, can be obtained by encoding its class description using the pre-trained text encoder $T_{\phi}$, as follows.
\begin{equation}
    \mathbf{w}_y = T_{\phi} (\text{``a paper of \{class name\}''}).
\end{equation}
The probability that a node $v$ belongs to class $y$ is thus
\begin{equation}
    p(y|v) = \textstyle \frac{\exp({\text{cos}(\mathbf{w}_y, \mathbf{v})})}{\sum_{y'=1}^N \exp({\text{cos}(\mathbf{w}_{y'}, \mathbf{v}))}},
\end{equation}
where the node embedding $\mathbf{v}$ is obtained via the pre-trained graph encoder $G_{\theta}$.

In the zero-shot setting, the prompts are discrete and are manually crafted. In the case of few-shot learning, where we assume a few labeled samples for each class, a prompt can be modeled as a sequence of $M+1$ continuous embeddings [$\mathbf{h}_1$, $\mathbf{h}_2$,\ldots,$\mathbf{h}_M$, $\mathbf{h}_\text{\{class\}}$], where $\mathbf{h}_1,\ldots,\mathbf{h}_M$ are trainable vectors and $\mathbf{h}_\text{\{class\}}$ is the embedding of the class name. The sequence length $M$ is a hyperparameter. The continuous embeddings can then be tuned using the few labeled samples as shown in Fig.~\ref{fig:intro}b. 



\begin{figure*}[t]
	\centering	\includegraphics[width=0.99\linewidth] {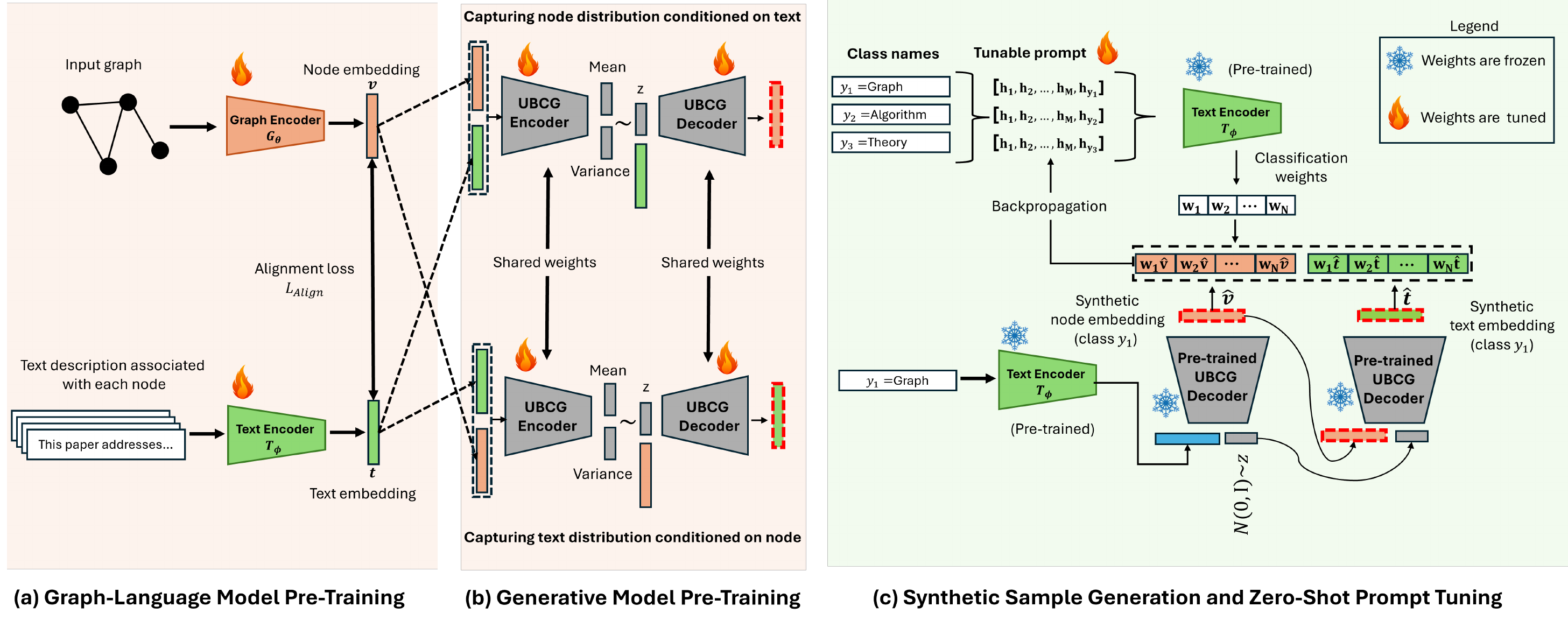} 
 \vspace{-5mm}	
 \caption{Overall framework of our zero-shot prompt tuning (ZPT). (a) and (b) are performed as pre-training steps, and (c) denotes the downstream classification phase. }
	\label{fig:training}
\end{figure*}

\section{Proposed Approach}
In this section, we explain the proposed ZPT approach in detail. We start with its overall framework, followed by  its key components.

\subsection{Overall Framework}

Our ZPT framework consists of three phases: the first two are pre-trained before seeing any downstream classification task, while the third addresses the downstream task. The overall framework is illustrated in Fig.~\ref{fig:training}.

In the first phase shown in Fig.~\ref{fig:training}a, we pre-train a graph-language model \cite{g2p2_wen2023augmenting} as described in Sect.~\ref{sub:g2p2-pretraining}. It jointly trains a graph encoder and a text encoder, aligning the graph and text representations to ensure semantic consistency between the two modalities. 

In the second phase  shown in Fig.~\ref{fig:training}b, the aligned node and text embeddings from the pre-trained encoders are utilized to train a conditional generative model, in order to generate class-specific synthetic samples for prompt tuning.  As motivated in Sect.~\ref{sec:intro}, the conditional generator must be universally applicable to all classification tasks, and capable of generating bimodal synthetic samples. Hence, we propose \CVAEfull\ (\CVAE), which learns the underlying node distribution conditioned on text, and the text distribution conditioned on nodes. Note that \CVAE\ can be pre-trained without any downstream task information, requiring neither labeled nodes nor class names. 

Moving to the downstream tasks, the third phase in Fig.~\ref{fig:training}c involves conditioning the pre-trained \CVAE\  with the class names to generate synthetic samples with both graph and text representations for each class. These generated synthetic samples are then used for continuous prompt tuning, not only improving adaptability to the zero-shot classes, but also enhancing robustness by eliminating the need for handcrafted discrete prompts.

\subsection{\CVAEfull}

To enable zero-shot prompt tuning without relying on manually designed discrete prompts, 
we propose \CVAEfull\ (\CVAE), a generative model to learn the underlying distributions in both graph and text modalities, each conditioned on the other modality, based on the architecture of Conditional Variational Autoencoder \cite{cvae_sohn2015learning}. As shown in Fig.~\ref{fig:training}b, \CVAE\ only needs to be pre-trained once, as it leverages only the node and text embeddings from the pre-trained encoders but not any downstream class information. 
Once trained, \CVAE\ can be conditioned on class names provided during downstream tasks to generate class-specific synthetic samples in the node embedding space. Subsequently, we further condition \CVAE\ on the synthetic  node embeddings to generate corresponding samples in the text embedding space.

\myparagraph{Capturing node  distribution.} The loss to capture the underlying distribution of nodes, conditioned on text, is formulated as follows.
\begin{align}
\mathcal{L}_{\text{gen}}^{\text{node}} &= -\mathbb{E}_{\mathbf{z} \sim q_{\zeta} (\cdot| \mathbf{v,t})} [\log p_{\psi} (\mathbf{v} | \mathbf{z,t})] + \mathtt{KL}(q_{\zeta} (\mathbf{z} | \mathbf{v,t}) \| p_{\psi} (\mathbf{z} | \mathbf{t})),\label{eq:vae_node}
\end{align}
where $\mathbf{v}$, $\mathbf{t}$ denote the node embedding and the corresponding text embedding from the pre-trained encoders, respectively, for a node $v$ on the graph. Furthermore, $\mathbf{z}$ is a latent variable, and $\mathtt{KL}$ represents the Kullback-Leibler (KL) divergence. Here, $\mathbf{t}$ serves as the condition vector, and $\mathbf{v}$ is the input node embedding whose distribution we aim to model. In particular, using $\mathbf{t}$ as the conditional input for node $v$ acts as a proxy for its class name, since $\mathbf{t}$ entails a semantically relevant representation of $v$'s class.

In this loss, the posterior $q_{\zeta} (\mathbf{z} | \mathbf{v,t})$ is modeled using an encoder network $E$ with parameters $\zeta$ which maps $\mathbf{v}$ from the input space to the latent space conditioned on $\mathbf{t}$. Similarly, $p_{\psi} (\mathbf{v} | \mathbf{z,t})$ can be modeled using a decoder network $D$ with parameters $\psi$, which maps from the latent space to the input $\mathbf{v}$ conditioned on $\mathbf{t}$. 
Specifically, the first term in the loss function can be seen as a reconstruction loss, while the second term  can be seen as minimizing the KL divergence between the predicted posterior $q_{\zeta} (\mathbf{z} | \mathbf{v,t})$ and the prior $p_{\psi} (\mathbf{z} | \mathbf{t})$.
The posterior is modeled as a multivariate normal distribution whose mean and variance are predicted by the encoder. The prior is drawn from a standard multivariate normal distribution $N(\mathbf{0},\mathbf{I})$.

\myparagraph{Capturing text distribution.} Similarly, the loss to capture the underlying distribution of text, conditioned on nodes, can be expressed as follows.
\begin{align}
\mathcal{L}_{\text{gen}}^{\text{text}} = -\mathbb{E}_{z\sim q_{\zeta} (\cdot | \mathbf{t,v})} [\log p_{\psi} (\mathbf{t} | \mathbf{z,v})] + \mathtt{KL}(q_{\zeta} (\mathbf{z} | \mathbf{t,v}) \| p_{\psi} (\mathbf{z} | \mathbf{v})).\label{eq:vae_text}
\end{align}
Different from Eq.~\eqref{eq:vae_node}, $\mathbf{v}$ becomes the condition vector and $\mathbf{t}$ is the input embedding whose distribution we aim to model. We model the posterior $q_{\zeta} (\mathbf{z} | \mathbf{t,v})$ and  $p_{\psi} (\mathbf{t} | \mathbf{z,v})$ using the same encoder $E$ and decoder $D$.

Hence, the overall loss of \CVAE\ is
\begin{align}
   \mathcal{L}_{\text{gen}} = \mathcal{L}_{\text{gen}}^{\text{node}} + \mathcal{L}_{\text{gen}}^{\text{text}}.
\end{align}

\subsection{Prompt Tuning Using Synthetic Samples}
In the downstream phase, consider a zero-shot node classification task as defined in Sect.~\ref{sec:setting}. To enable continuous prompt tuning, we utilize the pre-trained decoder $D$ of \CVAE\ to generate class-specific synthetic samples with bimodal embeddings. 

\myparagraph{Generation of synthetic node embeddings.} To generate samples for a given class, we first leverage its class name as the conditional input to \CVAE. Class names are the only available information and serve as textual descriptions that are semantically relevant to the class.
Specifically, we first sample the latent variable $\mathbf{z} \sim N(\mathbf{0},\mathbf{I})$ from a multivariate Normal distribution. At the same time, we pass the class name to the pre-trained text encoder and obtain its embedding. The latent variable $\mathbf{z}$ is then fed to the pre-trained decoder $D$ of \CVAE, conditioned on the class name embedding, to obtain the synthetic node embedding $\hat{\mathbf{v}}$ for the given class.

\myparagraph{Generation of synthetic text embeddings.} Once we obtain the synthetic node embeddings $\hat{\mathbf{v}}$, we proceed to generate the corresponding synthetic text embeddings $\hat{\mathbf{t}}$.
Specifically, we feed the same latent variable $\mathbf{z}$ used in the generation of $\hat{\mathbf{v}}$, along with the condition vector $\hat{\mathbf{v}}$, to the pre-trained decoder. 

\myparagraph{Prompt tuning.}
We formulate a continuous prompt as a sequence of continuous embeddings [$\mathbf{h}_1$, $\mathbf{h}_2$, \ldots,$\mathbf{h}_M$, $\mathbf{h}_{\text{\{class\}}}$] as outlined in Sect.~\ref{sub:g2p2-pretraining}.
The continuous prompt is passed to the pre-trained text encoder to obtain the classification weights for each class $y$:
\begin{align}
    \mathbf{w}_y = T_{\phi}([\mathbf{h}_1, \ldots,\mathbf{h}_M, \mathbf{h}_{\text{\{class\}}}]).
\end{align}
Then, the class probability is computed as 
\begin{align}
\label{Eq:weighted_prob}
    \hat{p}(y|\hat{\mathbf{v}},\hat{\mathbf{t}}) =
    \textstyle \lambda \frac{\exp(\text{cos}(\mathbf{w}_y, \hat{\mathbf{v}}))}{\sum_{y'=1}^N \exp(\text{cos}(\mathbf{w}_{y'}, \hat{\mathbf{v}}))} + (1-\lambda) \frac{\exp(\text{cos}(\mathbf{w}_y, \hat{\mathbf{t}}))}{\sum_{y'=1}^N \exp(\text{cos}(\mathbf{w}_{y'}, \hat{\mathbf{t}}))},
\end{align}
where the first term represents the probability based on the synthetic node embedding, and the second term represents the probability based on the synthetic text embedding. $\lambda \in [0,1]$ is a hyperparameter that balances the graph and text modalities. The prompts are then tuned using the standard cross-entropy loss.

\myparagraph{Inference.} Given a test node $v$ with node embedding $\mathbf{v}$ and text embedding $\mathbf{t}$, we compute the class-wise hybrid probability in the same way as Eq.~\eqref{Eq:weighted_prob}. Then, we predict the class with maximal probability.

\myparagraph{Computational overhead.} The complexity of the pre-training stage in G2P2 model is $O(|V|(\eta + \beta))$, where $|V|$ is the number of nodes, $\eta$ is the number of neighbors used in the prompt initialization, and $\beta$ is the batch size. In contrast, the overhead introduced by the proposed UBCG model is much smaller at $O(|V|/\beta)$. Empirically, the overall runtime on the Cora dataset for G2P2 is 1225s while for our approach is 1360s, incurring a negligible increase. Moreover, the proposed UBCG model has only 68,560 parameters, which are substantially fewer than the 63 million parameters of G2P2.






\section{Experiments}
In this section, we conduct experiments to evaluate our approach in comparison to various baselines, along with further model analysis. 

\begin{table}[t]
  \centering
  \small
  \caption{Statistical description of the datasets.}
  \vspace{-2mm}
    \begin{tabular}{l|r|r|r|r} \hline
    Datasets   & Cora  & Art   & Industial & MI \\ \hline
    No. nodes & 25,120 & 1,615,902 & 1,260,053 & 905,453 \\
    No. edges & 182,280 & 4,898,218 & 3,101,670 & 2,692,734 \\
    Avg. degree & 7.26  & 3.03  & 2.46  & 2.97 \\
    No. classes & 70    & 3,347 & 2,462 & 1,191 \\\hline
    \end{tabular}%
  \label{tab:dataset}%
\end{table}%

\begin{table*}[htbp]
  \centering
  \small
  \addtolength{\tabcolsep}{1mm}
  \caption{Zero-shot node classification performance of our proposed approach and the baselines. 
  The results are averaged across five runs. The results for Hound are obtained from prior work \cite{Hound_wang2024hound}.
  The best results are presented in bold. 
  }
  \vspace{-2mm}
    \begin{tabular}{l|cc|cc|cc|cc} \hline
     & \multicolumn{2}{c|}{Cora} & \multicolumn{2}{c|}{Art} & \multicolumn{2}{c|}{Industrial} & \multicolumn{2}{c}{MI} \\ 
      Method    & \multicolumn{1}{c}{Acc} & \multicolumn{1}{c|}{Macro F1} & \multicolumn{1}{c}{Acc} & \multicolumn{1}{c|}{Macro F1} & \multicolumn{1}{c}{Acc} & \multicolumn{1}{c|}{Macro F1} & \multicolumn{1}{c}{Acc} & \multicolumn{1}{c}{Macro F1} \\ \hline
    BERT  & 23.58±2.14 & 16.80±0.76 & 35.88±1.64 & 22.12±0.51 & 37.32±0.97 & 29.41±0.39 & 37.42±0.91 & 28.07±0.71 \\
    BERT* & 23.38±2.24 & 15.83±0.96 & 54.27±2.11 & 40.24±0.56 & 56.02±1.39 & 47.50±0.83 & 50.19±0.82 & 41.12±0.28 \\
    BERT* + d & 26.65±1.75 & 19.11±1.17 & 56.61±1.79 & 41.62±0.57 & 55.93±0.98 & 47.55±0.70 & 52.13±0.82 & 43.19±0.39 \\
    RoBERTa & 30.46±2.29 & 22.40±2.16 & 42.80±1.07 & 30.73±0.68 & 42.89±1.11 & 33.30±0.41 & 36.40±1.37 & 27.05±0.77 \\
    RoBERTa* & 39.58±1.43 & 32.83±1.26 & 34.77±0.74 & 21.98±0.16 & 37.78±0.36 & 27.85±0.30 & 32.17±0.78 & 23.09±0.28 \\
    RoBERTa* + d & 45.53±1.35 & 39.11±1.47 & 36.11±0.66 & 23.65±0.48 & 39.40±1.24 & 28.06±0.58 & 37.65±0.34 & 29.45±0.11 \\
    $\text{Llama3-8B}$ & 43.49±1.77 & 38.10±1.71 & 58.03±0.90 & 46.77±0.86 & 57.78±0.28 & 51.42±0.38 & 51.83±0.57 & 43.67±0.32 \\ \hline
    
    G2P2  & 63.52±2.94 & 58.42±1.84 & 76.52±0.60 & 64.71±0.38 & 76.66±0.32 & 69.59±0.31 & 74.60±0.64 & 67.13±0.71 \\
    G2P2 + d & 65.28±3.19 & 60.20±2.05 & 76.99±0.61 & 65.09±0.48 & 77.43±0.27 & 70.32±0.20 & 75.86±0.71 & 68.53±0.98 \\ 
    Hound + d & \textbf{69.21±1.35} & 61.41±1.82 & 78.22±1.70 & 67.71±0.02 & 81.99±0.58 & 73.84±0.33 & 79.85±1.35 & 72.58±0.79 \\\hline
    ZPT (Ours)  & 66.39±2.03 & 60.55±1.24 & 84.67±0.64 & 75.76±0.55 & 86.86±0.16 & \textbf{81.91±0.20} & 85.53±0.75 & 80.40±1.26 \\
    ZPT + Context (Ours) & 68.15±2.08 & \textbf{62.26±1.90} & \textbf{84.76±0.62} & \textbf{75.90±0.53} & \textbf{86.86±0.34} & 81.88±0.32 & \textbf{85.54±0.74} & \textbf{80.49±1.21} \\ \hline
    \end{tabular}%
  \label{tab:results_main}%
\end{table*}%

\begin{table*}[htbp]
  \centering
   \small
  \caption{Effect of synthetic sample generation and prompt tuning.}
  \vspace{-2mm}
    \begin{tabular}{c|c|c|cc|cc|cc|cc} \hline
          &       &    Downstream   & \multicolumn{2}{c|}{Cora} & \multicolumn{2}{c|}{Art } & \multicolumn{2}{c|}{Industial} & \multicolumn{2}{c}{MI} \\ 
    $\mathcal{L}^{\text{node}}_\text{gen}$ &  $\mathcal{L}^{\text{text}}_\text{gen}$ & training & \multicolumn{1}{c}{Acc} & \multicolumn{1}{c|}{Macro F1} & \multicolumn{1}{c}{Acc} & \multicolumn{1}{c|}{Macro F1} & \multicolumn{1}{c}{Acc} & \multicolumn{1}{c|}{Macro F1} & \multicolumn{1}{c}{Acc} & \multicolumn{1}{c}{Macro F1} \\ \hline
    \checkmark     & \checkmark     & Prompt Tuning & \textbf{68.15±2.08} & \textbf{62.26±1.90} & \textbf{84.76±0.62} & \textbf{75.90±0.53} & \textbf{86.86±0.34} & \textbf{81.88±0.32} & \textbf{85.54±0.74} & \textbf{80.49±1.21} \\
    \checkmark     & $\times$     & Prompt Tuning & 64.41±2.54 & 58.74±2.07 & 77.87±0.61 & 66.24±0.31 & 78.57±0.25 & 71.58±0.21 & 77.51±0.57 & 70.57±0.83 \\
    $\times$     & $\times$     & Discrete Prompt & 65.86±3.09 & 60.38±2.14 & 84.70±0.44 & 75.61±0.56 & 86.08±0.30 & 80.94±0.26 & 83.64±0.64 & 78.13±1.02 \\
    \checkmark     & \checkmark     & Simple Classifier & 63.81±3.23 & 58.66±2.25 & 81.76±0.81 & 71.99±0.39 & 84.76±0.39 & 79.37±0.18 & 83.52±0.63 & 77.84±1.08 \\
    \hline
    \end{tabular}%
  \label{tab:ablation_main}%
\end{table*}%

\subsection{Experimental Setup}
\subsubsection{Datasets}
The following four publicly available datasets, as summarized in Table ~\ref{tab:dataset}, are considered. Cora \cite{cora_mccallum00automating} is a citation network where the papers (nodes) are connected via the citation relations and the abstract serves as the textual description for each node. Each node is associated with a topic class. The remaining three datasets are derived from the Amazon reviews dataset \cite{ni2019justifying}, involving various categories including arts, crafts and sewing (Art), industrial and scientific (Industrial), and musical instruments (MI). Products and users are treated as nodes, with an edge between a user and a product if the user has reviewed it. Class labels correspond to product categories, and classification is performed only on product nodes. Additional details on the datasets and preprocessing are provided in prior work \cite{g2p2_wen2023augmenting}.

\subsubsection{Task and evaluation}
For our experiments, we follow the setup in earlier studies \cite{g2p2_wen2023augmenting, Hound_wang2024hound}. Specifically, we adopt a 5-way classification setting, where each task is formed by sampling five classes from the full set of available classes in the dataset. For our zero-shot learning scenario, each task contains only a query set without any labeled samples. Importantly, all classes in the dataset are covered across different tasks, though each individual task involves only a subset of five classes. During testing, we evaluate performance over multiple tasks to ensure comprehensive class coverage. The final results reported are the average performance across all tasks for each dataset.

\subsubsection{Baseline}
For our experiments, we consider various baselines broadly from two categories, namely, pre-trained language models and jointly pre-trained graph-language models. 

For  pre-trained language models, we compare to  \textbf{BERT} and \textbf{RoBERTa} models, which are pre-trained using masked language modeling. Furthermore, BERT and RoBERTa are further fine-tuned on our training data, denoted as \textbf{BERT*} and \textbf{RoBERTa*}, to mitigate the domain gap. These models perform zero-shot classification by comparing the cosine similarity of the embeddings of the textual description associated with each node to the embeddings of class names in the downstream task. Finally, instead of just using class names, we also provide the results of using discrete prompts such as ``a research paper of \{class name\}'', which is indicated by \textbf{+ d}. Additionally, we also compare with a large language model, \textbf{Llama3-8B}. 

For jointly pre-trained graph-language models, we consider \textbf{G2P2} and \textbf{Hound}. In particular, G2P2 first jointly pre-trains a graph encoder and a text encoder using alignment-based contrastive losses without any labeled samples. Then, during downstream tasks, it first obtains the class embeddings by passing class names to the text encoder. The test nodes in the query set are then classified by comparing the cosine similarity between the node embedings produced by the graph encoder and the class embeddings. In contrast, Hound introduces more self-supervision signals, including node perturbation, text matching, and semantic negation, enabling the joint pre-training of an additional negative text encoder. During downstream tasks, class names are passed to the text encoder to obtain positive class embeddings, and to the negative text encoder to obtain negative class embeddings. These embeddings are used to compute positive and negative probabilities for each test node embedding produced by the graph encoder. The positive probability and the complement of the negative probability are then fused to produce the final prediction.

\subsubsection{Implementation Details}
For our approach ZPT, we present two variants, based on how the synthetic samples are generated. First, \textbf{ZPT} denotes the variant where the synthetic samples are generated using just the class name as the conditional input. Second, \textbf{ZPT + Context} denotes the variant in which additional contextual words like ``a paper of \{class name\}'' are used as the conditional input, similar to the handcrafted discrete prompts used in G2P2. 
Full implementation details of the proposed approach and the baselines are provided in Appendices~\ref{app:impl_details:proposed} and ~\ref{app:impl_details:baselines}.


\subsection{Results}
We present the results of various baseline models and our proposed ZPT approach in Table \ref{tab:results_main}. We make several observations.

First, using only language models like BERT, RoBERTa, and Llama-3-8B results in inferior performance. This is because these models rely solely on textual descriptions associated with each node, without leveraging the underlying graph structure. In contrast, graph-language models such as G2P2 and Hound perform better than pure language models by incorporating the graph structure. 
In particular, Hound improves zero-shot predictions over G2P2 by incorporating additional self-supervision and utilizing both positive and negative text encoders. 

Second, our proposed ZPT approach consistently achieves superior performance by leveraging synthetic samples for prompt tuning. Although it trails Hound on the Cora dataset in terms of accuracy, it outperforms Hound in macro-F1 score. Notably, macro-F1 is a more appropriate metric than accuracy for Cora due to its class imbalance.

Third, models using discrete prompts such as ``a paper of \{class name\}'' (denoted by + d) generally outperform those relying solely on class names. 
Similar to the discrete prompts, incorporating contextual words like ``a paper of \{class name\}'' into the conditioning input for ZPT (denoted by + Context), rather than using only class names, often leads to further improvements. While the best performance among several handcrafted discrete prompts or contextual words is reported here, selecting the optimal prompts or contextual words requires significant manual engineering and often lacks a principled strategy. A performance comparison using different prompts or contextual words is provided in Appendix~\ref{app:prompts}. Nevertheless, it is worth noting that the vanilla ZPT, without any handcrafted contextual words, remains competitive. 

\begin{figure}[tbp]
	\centering
	\includegraphics[scale=0.35]{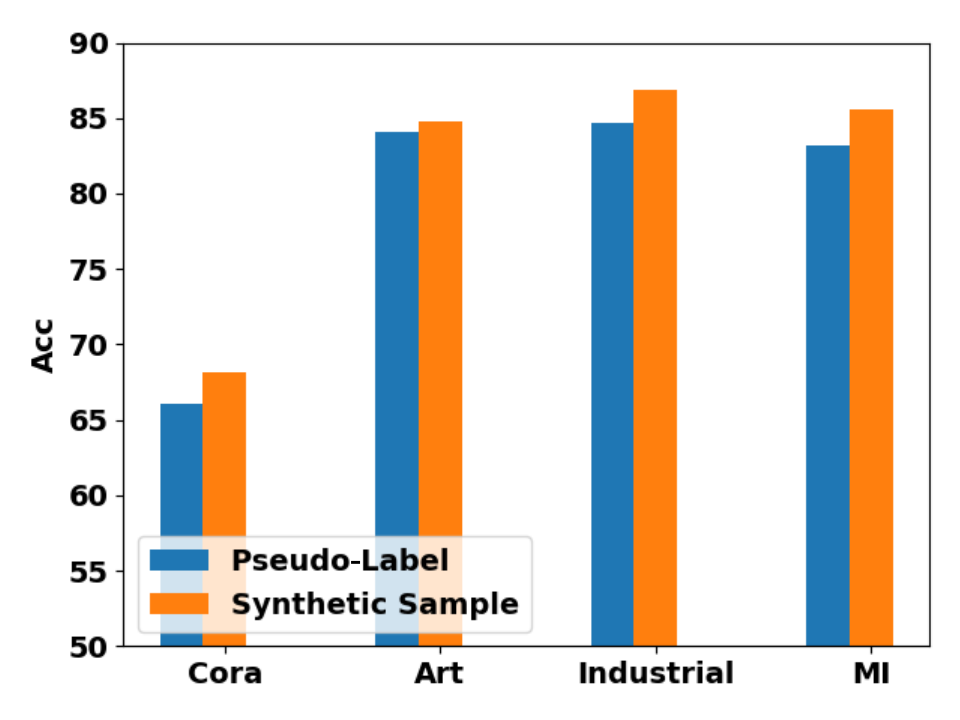}
         \vspace{-2mm}
	\caption{Performance of pseudo-labels vs.~synthetic samples.}
	\label{fig:pseudo_label}
\end{figure}

\begin{figure*}[tbp]
\centering

    \begin{subfigure}[t]{0.33\textwidth}
        {\includegraphics[width=\textwidth]{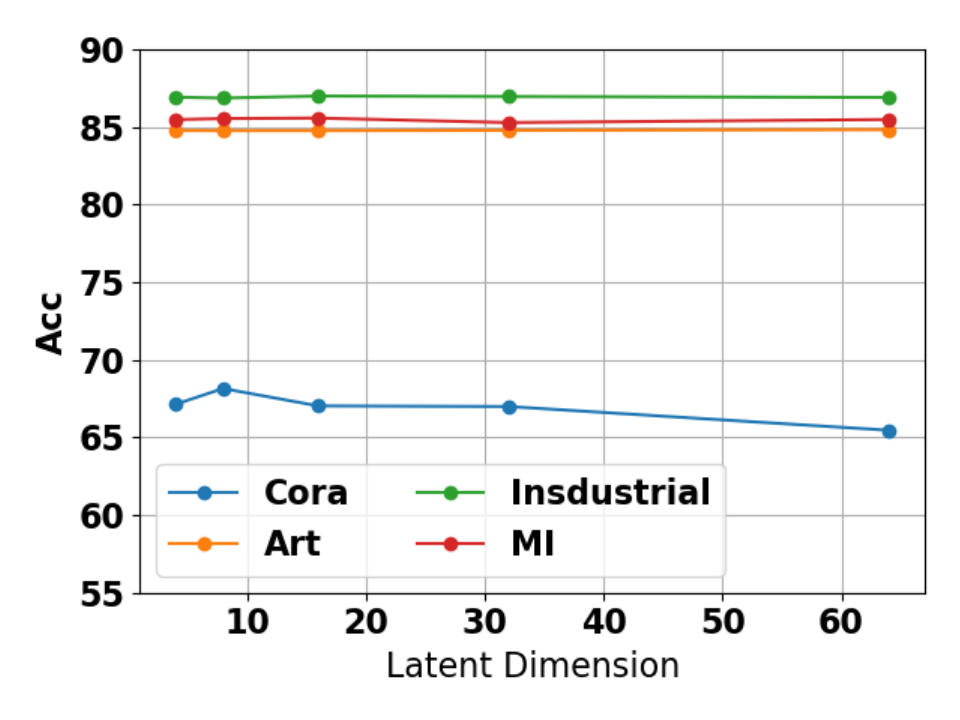}}
        \vspace{-6mm}
        \caption{Effect of Latent Dimension}
    \end{subfigure}
    \begin{subfigure}[t]{0.33\textwidth}
        {\includegraphics[width=\textwidth]{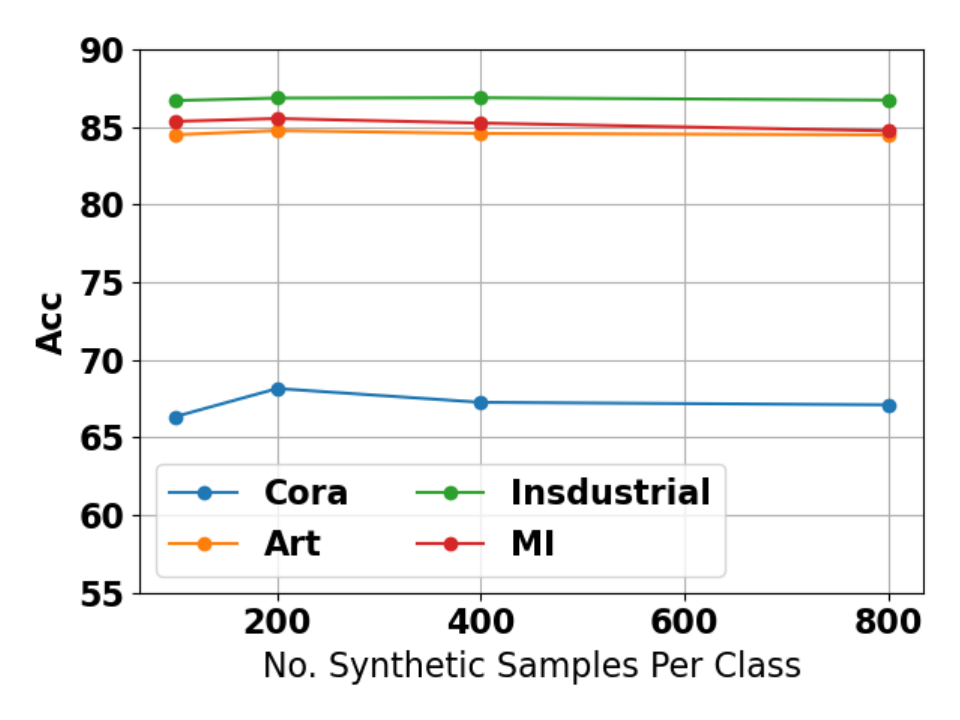}}
        \vspace{-6mm}
        \caption{Effect of Synthetic Samples}
    \end{subfigure}
     \begin{subfigure}[t]{0.33\textwidth}
        {\includegraphics[width=\textwidth]{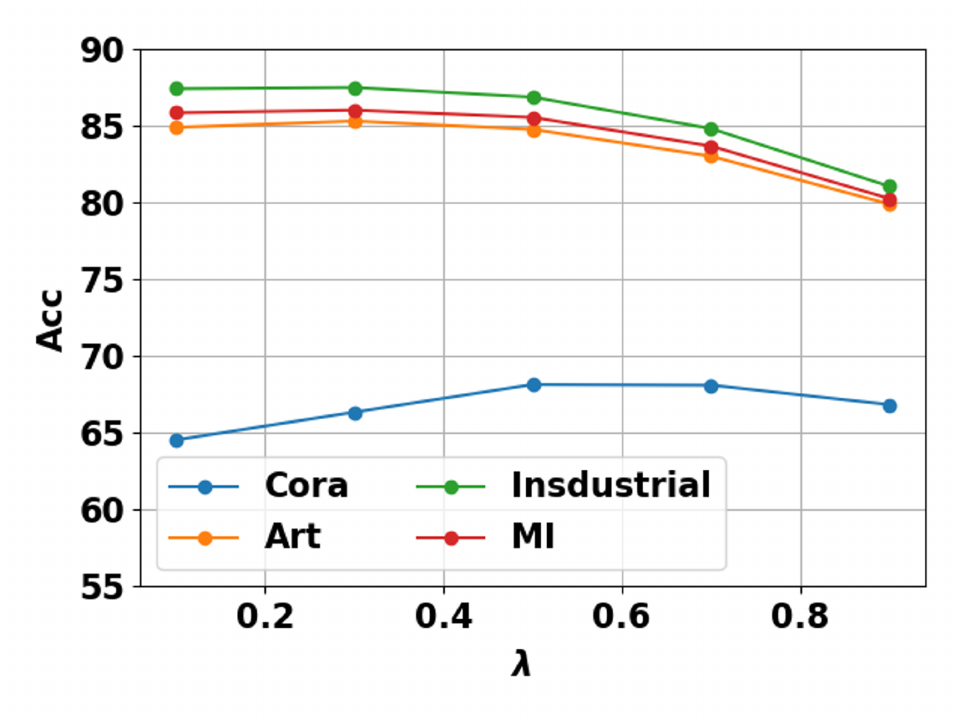}}
        \vspace{-6mm}
        \caption{Effect of $\lambda$}
    \end{subfigure}
 \vspace{-2mm}
\caption{Sensitivity analysis of model parameters.}
\label{fig:Ablation}
\end{figure*}

\subsection{Ablation Study}\label{sec:expt:ablation}
We present a series of ablation studies to evaluate the impact of different components on the overall performance of our model. Unless otherwise stated, we utilize discrete prompts (+ d) or contextual words (+ Context) in the following experiments. 

\myparagraph{Effect of synthetic sample generation and prompt tuning.} Table~\ref{tab:ablation_main} shows the effect of prompt tuning using the generated synthetic samples. Specifically, we investigate the following questions.

(1) \emph{What is the effect of training the \CVAE\ model to generate both the node and the corresponding text embeddings?} By comparing the first two rows, it can be seen that training the \CVAE\ model to generate embeddings for both modalities outperforms the variant that generates just the node embeddings. 

(2)  \emph{What is the effect of prompt tuning as compared to just using discrete prompts for zero-shot node classification?} By comparing the first and third rows, it can be seen that prompt tuning using synthetic samples consistently gives better performance than just using discrete prompts. 

(3) \emph{What is the effect of prompt tuning as compared to training a classifier directly on the generated synthetic samples?} By comparing the first and last rows, it can be seen that training a simple classifier on the generated synthetic samples significantly degrades  performance compared to prompt tuning.

\myparagraph{Pseudo-labels vs. synthetic samples.} As discussed in Sect.~\ref{sec:intro}, a na\"ive  workaround for prompt tuning in the zero-shot node classification setting is to use existing zero-shot approaches, such as discrete prompt-based methods, and assign pseudo-labels to the unlabeled nodes in downstream tasks. Hence, in this experiment, we compare the performance of the na\"ive approach, which assigns pseudo-labels using the G2P2 zero-shot model, and the proposed ZPT approach. The results are presented in Fig.~\ref{fig:pseudo_label}. We observe that our proposed ZPT approach outperforms the  na\"ive approach, which further highlights the advantages of synthetic samples. Implementation details of the pseudo-labeling approach are provided in Appendix~\ref{app: Implementation Pseudo Label}.

\myparagraph{Effect of different embeddings in downstream tasks.} 
During downstream prompt tuning and inference, we utilize both the node embeddings and its corresponding text embeddings as shown in Eq.~\eqref{Eq:weighted_prob}. 
In this experiment, we demonstrate that fusing both modalities in downstream tasks is beneficial. In Table~\ref{tab:node_text_hybrid}, we compare several variants of G2P2 and ZPT  utilizing only the node embeddings (*-Node), only the text embeddings (*-Text), and the hybrid approach that fuses both modalities (*-Hybrid).  For both G2P2 and ZPT, while the relative performance of node embeddings and text embeddings varies across datasets, the hybrid approach generally demonstrates robust performance across different datasets. Furthermore, our proposed ZPT approach outperforms G2P2 in most cases across the three variants.
Hence, for ZPT, the hybrid approach offers a reliable means to leverage both node and text information in downstream tasks.

\begin{table}[tbp]
  \centering
  \caption{Accuracy comparison of G2P2 and ZPT using different embeddings in downstream tasks. }
  \vspace{-2mm}
  \small
    \begin{tabular}{l|c|c|c|c}\hline
          & Cora  & Art   & Industrial & MI \\ \hline
    G2P2-Node & 65.28±3.19 & 76.99±0.61 & 77.43±0.27 & 75.86±0.71 \\
    ZPT-Node & \textbf{66.39±2.35} & \textbf{77.86±0.57} & \textbf{78.46±0.25} & \textbf{77.9±0.33} \\\hline
          
    G2P2-Text & 61.00±2.19 & \textbf{84.99±0.48} & 86.68±0.26 & 83.78±0.53 \\
    ZPT-Text & \textbf{63.01±1.07} & 84.45±0.59 & \textbf{87.20±0.27} & \textbf{85.50±0.71} \\\hline
         
    G2P2-Hybrid & 65.86±3.09 & 84.70±0.44 & 86.08±0.30 & 83.64±0.64 \\
    ZPT-Hybrid & \textbf{68.15±2.08} & \textbf{84.76±0.62} & \textbf{86.86±0.34} & \textbf{85.54±0.74} \\ \hline
    \end{tabular}%
  \label{tab:node_text_hybrid}%
\end{table}%

\subsection{Sensitivity Analysis}
We investigate the impact of various hyperparameters on the performance of our proposed approach. 


\myparagraph{Latent dimension in \CVAE.} The dimension of the latent variable $\mathbf{z}$ in \CVAE\   denotes the amount of compressibility of the encoder. We vary the latent dimension size over $\{4, 8, 16, 32, 64\}$ and present the results in Fig.~\ref{fig:Ablation}a. The model is generally robust to different values of the latent dimension. Overall, we use a latent dimension of 8 across all our datasets.

\myparagraph{Number of synthetic samples.} Using the trained \CVAE, we can generate as many samples as required in downstream tasks. Hence, we investigate the effect of number of synthetic samples used for prompt tuning. Specifically, we vary the number of synthetic samples for each class over $\{100, 200, 400, 800\}$ and present the results in Fig.~\ref{fig:Ablation}b. Overall, generating 200 synthetic samples for each  class gives better performance across all four datasets.

\myparagraph{Balance parameter $\lambda$.} In the hybrid strategy,  $\lambda$ is used to balance the trade-off between the node and text embeddings. Here, we investigate the impact of $\lambda$ on the overall performance. Specifically, we vary  $\lambda$ over $\{0.1, 0.3, 0.5, 0.7, 0.9\}$ and report the results in Fig.~\ref{fig:Ablation}c. 
The performance trends on Cora and the Amazon datasets differ, reflecting dataset-specific characteristics and differing emphasis on the two modalities. Nevertheless, setting $\lambda = 0.5$, which assigns equal weight to both modalities, provides a robust choice.

\subsection{Visualization of Synthetic Samples}

\begin{figure}
	\centering
	\includegraphics[width=0.99\linewidth]{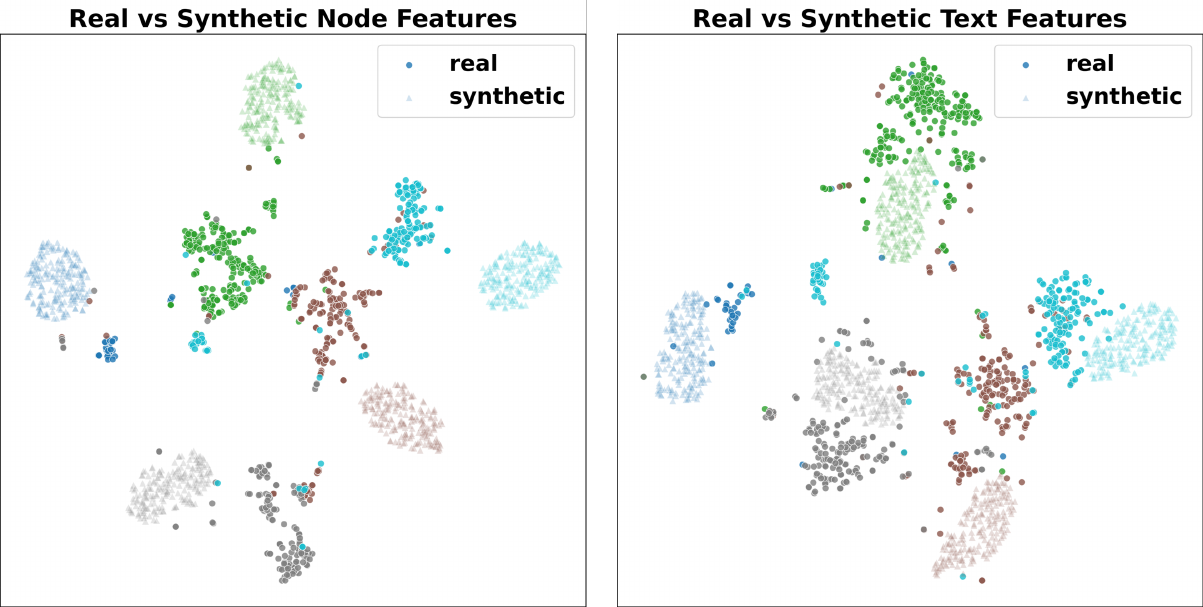}
	\caption{$t$-SNE visualization of real and synthetic samples, as represented by dark and faded markers, respectively. }
		\label{fig:t-sne visualization}
\end{figure}

Finally, we examine the relationship between real and synthetic samples through a visualization. In Fig.~\ref{fig:t-sne visualization}, we present the $t$-SNE embeddings of real and synthetic samples for five classes on the Cora dataset. In both node and text feature spaces, the synthetic samples form crisp clusters that align with the real data distribution, indicating that the \CVAE\ model captures key class-specific structures. However, we observe a small offset between each synthetic cluster and its corresponding real cluster, as \CVAE\ was trained without using any class information or labeled nodes for conditioning.

\section{Conclusion}

In this work, we address the challenge of zero-shot node classification on text-attributed graphs by introducing the Zero-shot Prompt Tuning (ZPT) approach. By utilizing a \CVAEfull \space (\CVAE) that synthesizes both graph and text representations based on class names, we effectively eliminate the need for labeled nodes or pseudo-labels for prompt tuning. Our generative model, trained once during pre-training, is universally applicable to novel downstream tasks, ensuring scalability and adaptability. Extensive experiments validate the efficacy of the proposed ZPT approach in zero-shot node classification. 

\section*{Acknowledgment}

This research is supported by the Ministry of Education, Singapore, under its Academic Research Fund Tier 1 grant (22-SIS-SMU-054). 

\appendix
\section*{Appendices}


 \section{Implementation Details}\label{app:impl_details}
\subsection{Proposed Approach}\label{app:impl_details:proposed}
 For our graph-language model, we follow prior work~\cite{g2p2_wen2023augmenting}. Specifically, the text encoder is a 12-layer, 512-wide Transformer model with 8 attention heads. The maximum sequence length is capped at 128. The graph encoder is a GCN with two layers, each with 128 dimensions and LeakyReLU activation. The graph-text pre-training starts from scratch with randomly initialized weights. The co-efficient hyperparameter for the summary-based contrastive losses is set to $0.1$ on Cora and $10$ on the three Amazon review datasets. We use the Adam optimizer with a learning rate of $2\times10^{-5}$ for training the model. For \CVAE, the encoder is modeled using a multi-layer perceptron (MLP) with two hidden layers. The input dimension is 256 (input embedding concatenated with condition embedding) and the hidden layers dimension is 128. The output is  the mean and variance in the latent space with a dimension of 8. Similarly, the decoder is modeled using MLP with one hidden layer. The hidden layer dimension is 64 followed by the output of dimension 128. The input is the sampled latent vector concatenated with the condition embedding. All layers use ReLU activation. \CVAE\  is trained with a learning rate of $10^{-3}$ using the Adam optimizer. For the downstream tasks, we generate 200 synthetic node and text embeddings per class and use them to tune the prompts. The fusion parameter $\lambda$ is set to 0.5 for all datasets. 

\subsection{Baselines}\label{app:impl_details:baselines}
We follow the settings in prior work \cite{g2p2_wen2023augmenting, Hound_wang2024hound} for the pre-trained language models, namely, BERT and RoBERTa. Specifically, the models and the implementation of masked language modeling are based on Hugging Face's Transformers library. Furthermore, we also consider a pre-trained Llama3-8B model as a baseline. In general, pre-trained language models perform zero-shot classification by first obtaining the class embedding using the provided class names and the text embedding associated with the target node. Then, they assign the class whose class embedding is the closest to the target text embedding based on cosine similarity. However, for Llama3-8B, we observe that using a prompt of the form ``This sentence: a \{text\} means in one word:'', where \{text\} is either a class name or the associated textual description of the target node, and taking the last hidden state as the embedding yields better classification performance. For G2P2, we employ the authors' code for reproducing the results. Furthermore, the G2P2-Hybrid variant in Sect.~\ref{sec:expt:ablation} adopts the same fusion strategy as Eq.~\eqref{Eq:weighted_prob} and sets the fusion parameter $\lambda$ to 0.5, consistent with our proposed approach. For Hound, we report results from the original paper \cite{Hound_wang2024hound}.

\subsection{Prompt Tuning Using Pseudo-labels}
\label{app: Implementation Pseudo Label}
We first apply G2P2 to assign pseudo-labels to all nodes in the graph in a zero-shot manner using a discrete prompt over all classes in the dataset. The discrete prompts achieving the best performance on each dataset are chosen, consistent with those used by G2P2 + d in Table~\ref{tab:results_main}.
For each class in a downstream task, we then sample up to 200 nodes labeled as that class and use them for prompt tuning. If fewer than 200 pseudo-labeled samples are available for a class, all available samples are used instead. The prompts are tuned for one epoch with a batch size of 64, 
using a learning rate of $2\times 10^{-5}$.

 \section{Effect of Discrete Prompts and Contexts}
 \label{app:prompts}

We investigate the effect of different discrete prompts or contextual words as used by G2P2 and ZPT, respectively, in Table~\ref{tab:comparison}. The performance of different prompts or contexts varies significantly; hence, they often require substantial manual effort to design. Moreover, the results demonstrate the robustness of ZPT, which consistently outperforms G2P2 across various prompts and contexts. These findings also suggest that our proposed \CVAE\ model can effectively leverage such semantic contexts to generate synthetic samples.

\begin{table}[tbp]
    \centering
    \small
    \caption{Accuracy comparison of G2P2 and ZPT across different prompts or contextual words on the Cora dataset.}
    \vspace{-2mm}
    \begin{tabular}{l|c|c}
        \hline
        \textbf{Prompts} & \textbf{G2P2} & \textbf{ZPT} \\
        \hline  
        \{Class Name\} & 64.06$\pm$2.61 & 66.39$\pm$2.03 \\
        a \{Class Name\} & 65.86$\pm$3.09 & 68.15$\pm$2.08 \\
        an \{Class Name\} & 65.21$\pm$3.00 & 67.82$\pm$1.51 \\
        a paper of \{Class Name\} & 62.94$\pm$3.10 & 65.84$\pm$2.25 \\
        a research of \{Class Name\} & 61.68$\pm$2.31 & 65.10$\pm$1.41 \\
        a research paper of \{Class Name\} & 62.02$\pm$1.77 & 63.87$\pm$2.06 \\
        \hline 
    \end{tabular}
    \label{tab:comparison}
    \vspace{-3mm}
\end{table}

\section{Alternative Pre-trained Text Encoder}

In Table~\ref{tab:sbert}, we explore the effect of replacing the simple Transformer-based text encoder with a pre-trained SentenceTransformer (SBERT) \cite{reimers-2019-sentence-bert} on the Cora dataset. G2P2 with SBERT as its text encoder outperforms SBERT alone, as SBERT does not consider the graph structure. Furthermore, our proposed ZPT model with SBERT as its text encoder achieves the best performance.

\begin{table}[htbp]
  \centering
  \small
  \caption{Effect of using SBERT as the text encoder on Cora.}
  \vspace{-2mm}
    \begin{tabular}{l|c|c} \hline
          & Acc   & Macro F1 \\ \hline 
    SBERT & 77.86±1.61 & 73.29±1.88 \\
    G2P2-SBERT & 80.19±1.83 & 75.34±2.04 \\
    ZPT-SBERT & \textbf{81.73±1.79} & \textbf{77.71±1.43} \\ \hline
    \end{tabular}%
  \label{tab:sbert}%
\end{table}%

\clearpage
\section*{Ethical Considerations}
Our work proposes a label-free prompt tuning approach for zero-shot node classification on text-attributed graphs. From a social perspective, reducing reliance on labeled data can broaden access to machine learning models in low-resource settings, potentially benefiting underrepresented communities where annotated data and domain expertise are limited. While our work does not raise immediate or specific ethical concerns, the broader application of graph machine learning and artificial intelligence techniques may still entail inherent risks.


\balance
\bibliographystyle{ACM-Reference-Format}
\bibliography{sample-base}

@String{Computing = "Computing" }

@String{Computer = "{IEEE} Computer" }

@String{Academic = "Academic Press" }

@String{Chelsea = "Chelsea" }

@ArtifactSoftware{R,
    title = {R: A Language and Environment for Statistical Computing},
    author = {{R Core Team}},
    organization = {R Foundation for Statistical Computing},
    address = {Vienna, Austria},
    year = {2019},
    url = {https://www.R-project.org/},
}

@article{GCN_kipf2016semi,
  title={Semi-supervised classification with graph convolutional networks},
  author={Kipf, Thomas N and Welling, Max},
  journal={arXiv preprint arXiv:1609.02907},
  year={2016}
}

@article{GAT_velivckovic2017graph,
  title={Graph attention networks},
  author={Veli{\v{c}}kovi{\'c}, Petar and Cucurull, Guillem and Casanova, Arantxa and Romero, Adriana and Lio, Pietro and Bengio, Yoshua},
  journal={arXiv preprint arXiv:1710.10903},
  year={2017}
}

@article{variational_graph_autoencoders_kipf2016variational,
  title={Variational graph auto-encoders},
  author={Kipf, Thomas N and Welling, Max},
  journal={arXiv preprint arXiv:1611.07308},
  year={2016}
}

@inproceedings{GIN_xu2018powerful,
  title={How Powerful are Graph Neural Networks?},
  author={Xu, Keyulu and Hu, Weihua and Leskovec, Jure and Jegelka, Stefanie},
  booktitle={International Conference on Learning Representations},
  year={2018}
}

@inproceedings{DGI_velivckovicdeep,
  title={Deep Graph Infomax},
  author={Veli{\v{c}}kovi{\'c}, Petar and Fedus, William and Hamilton, William L and Li{\`o}, Pietro and Bengio, Yoshua and Hjelm, R Devon},
  booktitle={International Conference on Learning Representations}
}

@article{graphcl_you2020graph,
  title={Graph contrastive learning with augmentations},
  author={You, Yuning and Chen, Tianlong and Sui, Yongduo and Chen, Ting and Wang, Zhangyang and Shen, Yang},
  journal={Advances in neural information processing systems},
  volume={33},
  pages={5812--5823},
  year={2020}
}

@inproceedings{graphmae_hou2022graphmae,
  title={Graphmae: Self-supervised masked graph autoencoders},
  author={Hou, Zhenyu and Liu, Xiao and Cen, Yukuo and Dong, Yuxiao and Yang, Hongxia and Wang, Chunjie and Tang, Jie},
  booktitle={Proceedings of the 28th ACM SIGKDD Conference on Knowledge Discovery and Data Mining},
  pages={594--604},
  year={2022}
}

@inproceedings{GCA_zhu2021graph,
  title={Graph contrastive learning with adaptive augmentation},
  author={Zhu, Yanqiao and Xu, Yichen and Yu, Feng and Liu, Qiang and Wu, Shu and Wang, Liang},
  booktitle={Proceedings of the web conference 2021},
  pages={2069--2080},
  year={2021}
}

@inproceedings{graphmae2_hou2023graphmae2,
  title={Graphmae2: A decoding-enhanced masked self-supervised graph learner},
  author={Hou, Zhenyu and He, Yufei and Cen, Yukuo and Liu, Xiao and Dong, Yuxiao and Kharlamov, Evgeny and Tang, Jie},
  booktitle={Proceedings of the ACM web conference 2023},
  pages={737--746},
  year={2023}
}

@inproceedings{mgae_wang2017mgae,
  title={Mgae: Marginalized graph autoencoder for graph clustering},
  author={Wang, Chun and Pan, Shirui and Long, Guodong and Zhu, Xingquan and Jiang, Jing},
  booktitle={Proceedings of the 2017 ACM on Conference on Information and Knowledge Management},
  pages={889--898},
  year={2017}
}

@article{universl_prompt_fang2024universal,
  title={Universal prompt tuning for graph neural networks},
  author={Fang, Taoran and Zhang, Yunchao and Yang, Yang and Wang, Chunping and Chen, Lei},
  journal={Advances in Neural Information Processing Systems},
  volume={36},
  year={2024}
}

@inproceedings{liu2023graphprompt,
  title={Graphprompt: Unifying pre-training and downstream tasks for graph neural networks},
  author={Liu, Zemin and Yu, Xingtong and Fang, Yuan and Zhang, Xinming},
  booktitle={Proceedings of the ACM Web Conference 2023},
  pages={417--428},
  year={2023}
}

@inproceedings{sun2022gppt,
  title={Gppt: Graph pre-training and prompt tuning to generalize graph neural networks},
  author={Sun, Mingchen and Zhou, Kaixiong and He, Xin and Wang, Ying and Wang, Xin},
  booktitle={Proceedings of the 28th ACM SIGKDD Conference on Knowledge Discovery and Data Mining},
  pages={1717--1727},
  year={2022}
}

@inproceedings{graph_few_shot_yao2020graph,
  title={Graph few-shot learning via knowledge transfer},
  author={Yao, Huaxiu and Zhang, Chuxu and Wei, Ying and Jiang, Meng and Wang, Suhang and Huang, Junzhou and Chawla, Nitesh and Li, Zhenhui},
  booktitle={Proceedings of the AAAI conference on artificial intelligence},
  volume={34},
  number={04},
  pages={6656--6663},
  year={2020}
}

@inproceedings{Graph_prototypical_few_shot_ding2020graph,
  title={Graph prototypical networks for few-shot learning on attributed networks},
  author={Ding, Kaize and Wang, Jianling and Li, Jundong and Shu, Kai and Liu, Chenghao and Liu, Huan},
  booktitle={Proceedings of the 29th ACM International Conference on Information \& Knowledge Management},
  pages={295--304},
  year={2020}
}

@inproceedings{Meta-gnn_zhou2019meta,
  title={Meta-gnn: On few-shot node classification in graph meta-learning},
  author={Zhou, Fan and Cao, Chengtai and Zhang, Kunpeng and Trajcevski, Goce and Zhong, Ting and Geng, Ji},
  booktitle={Proceedings of the 28th ACM International Conference on Information and Knowledge Management},
  pages={2357--2360},
  year={2019}
}

@inproceedings{all_in_one_sun2023all,
  title={All in one: Multi-task prompting for graph neural networks},
  author={Sun, Xiangguo and Cheng, Hong and Li, Jia and Liu, Bo and Guan, Jihong},
  booktitle={Proceedings of the 29th ACM SIGKDD Conference on Knowledge Discovery and Data Mining},
  pages={2120--2131},
  year={2023}
}

@inproceedings{kenton2019bert,
  title={Bert: Pre-training of deep bidirectional transformers for language understanding},
  author={Kenton, Jacob Devlin Ming-Wei Chang and Toutanova, Lee Kristina},
  booktitle={Proceedings of naacL-HLT},
  volume={1},
  pages={2},
  year={2019},
  organization={Minneapolis, Minnesota}
}

@article{radford2018improving,
  title={Improving language understanding by generative pre-training},
  author={Radford, Alec},
  year={2018}
}

@article{raffel2020exploring,
  title={Exploring the limits of transfer learning with a unified text-to-text transformer},
  author={Raffel, Colin and Shazeer, Noam and Roberts, Adam and Lee, Katherine and Narang, Sharan and Matena, Michael and Zhou, Yanqi and Li, Wei and Liu, Peter J},
  journal={Journal of machine learning research},
  volume={21},
  number={140},
  pages={1--67},
  year={2020}
}

@article{gpt3_brown2020language,
  title={Language models are few-shot learners},
  author={Brown, Tom B},
  journal={arXiv preprint arXiv:2005.14165},
  year={2020}
}

@inproceedings{prompt_tuning_wang2022towards,
  title={Towards Unified Prompt Tuning for Few-shot Text Classification},
  author={Wang, Jianing and Wang, Chengyu and Luo, Fuli and Tan, Chuanqi and Qiu, Minghui and Yang, Fei and Shi, Qiuhui and Huang, Songfang and Gao, Ming},
  booktitle={Findings of the Association for Computational Linguistics: EMNLP 2022},
  pages={524--536},
  year={2022}
}

@inproceedings{hu2022knowledgeable,
  title={Knowledgeable Prompt-tuning: Incorporating Knowledge into Prompt Verbalizer for Text Classification},
  author={Hu, Shengding and Ding, Ning and Wang, Huadong and Liu, Zhiyuan and Wang, Jingang and Li, Juanzi and Wu, Wei and Sun, Maosong},
  booktitle={Proceedings of the 60th Annual Meeting of the Association for Computational Linguistics (Volume 1: Long Papers)},
  pages={2225--2240},
  year={2022}
}

@inproceedings{zhong2021adapting,
  title={Adapting Language Models for Zero-shot Learning by Meta-tuning on Dataset and Prompt Collections},
  author={Zhong, Ruiqi and Lee, Kristy and Zhang, Zheng and Klein, Dan},
  booktitle={Findings of the Association for Computational Linguistics: EMNLP 2021},
  pages={2856--2878},
  year={2021}
}

@article{yang2019xlnet,
  title={XLNet: Generalized Autoregressive Pretraining for Language Understanding},
  author={Yang, Zhilin},
  journal={arXiv preprint arXiv:1906.08237},
  year={2019}
}

@article{sun2019ernie,
  title={Ernie: Enhanced representation through knowledge integration},
  author={Sun, Yu and Wang, Shuohuan and Li, Yukun and Feng, Shikun and Chen, Xuyi and Zhang, Han and Tian, Xin and Zhu, Danxiang and Tian, Hao and Wu, Hua},
  journal={arXiv preprint arXiv:1904.09223},
  year={2019}
}

@inproceedings{DGPN_wang2021zero,
  title={Zero-shot node classification with decomposed graph prototype network},
  author={Wang, Zheng and Wang, Jialong and Guo, Yuchen and Gong, Zhiguo},
  booktitle={Proceedings of the 27th ACM SIGKDD conference on knowledge discovery \& data mining},
  pages={1769--1779},
  year={2021}
}

@inproceedings{yue2022dual,
  title={Dual bidirectional graph convolutional networks for zero-shot node classification},
  author={Yue, Qin and Liang, Jiye and Cui, Junbiao and Bai, Liang},
  booktitle={Proceedings of the 28th acm sigkdd conference on knowledge discovery and data mining},
  pages={2408--2417},
  year={2022}
}

@article{ju2023zero,
  title={Zero-shot node classification with graph contrastive embedding network},
  author={Ju, Wei and Qin, Yifang and Yi, Siyu and Mao, Zhengyang and Zheng, Kangjie and Liu, Luchen and Luo, Xiao and Zhang, Ming},
  journal={Transactions on Machine Learning Research},
  year={2023}
}

@inproceedings{one_for_all_liuone,
  title={One For All: Towards Training One Graph Model For All Classification Tasks},
  author={Liu, Hao and Feng, Jiarui and Kong, Lecheng and Liang, Ningyue and Tao, Dacheng and Chen, Yixin and Zhang, Muhan},
  booktitle={The Twelfth International Conference on Learning Representations}
}

@inproceedings{graphgpt_tang2024graphgpt,
  title={Graphgpt: Graph instruction tuning for large language models},
  author={Tang, Jiabin and Yang, Yuhao and Wei, Wei and Shi, Lei and Su, Lixin and Cheng, Suqi and Yin, Dawei and Huang, Chao},
  booktitle={Proceedings of the 47th International ACM SIGIR Conference on Research and Development in Information Retrieval},
  pages={491--500},
  year={2024}
}

@inproceedings{g2p2_wen2023augmenting,
  title={Augmenting low-resource text classification with graph-grounded pre-training and prompting},
  author={Wen, Zhihao and Fang, Yuan},
  booktitle={Proceedings of the 46th International ACM SIGIR Conference on Research and Development in Information Retrieval},
  pages={506--516},
  year={2023}
}

@article{Hound_wang2024hound,
  title={Hound: Hunting Supervision Signals for Few and Zero Shot Node Classification on Text-attributed Graph},
  author={Wang, Yuxiang and Yan, Xiao and Jin, Shiyu and Xu, Quanqing and Yang, Chuanhui and Zhu, Yuanyuan and Hu, Chuang and Du, Bo and Jiang, Jiawei},
  journal={arXiv preprint arXiv:2409.00727},
  year={2024}
}

@inproceedings{web_scale_recommender_systemsying2018graph,
  title={Graph convolutional neural networks for web-scale recommender systems},
  author={Ying, Rex and He, Ruining and Chen, Kaifeng and Eksombatchai, Pong and Hamilton, William L and Leskovec, Jure},
  booktitle={Proceedings of the 24th ACM SIGKDD international conference on knowledge discovery \& data mining},
  pages={974--983},
  year={2018}
}

@inproceedings{zhang2022robust,
  title={Robust self-supervised structural graph neural network for social network prediction},
  author={Zhang, Yanfu and Gao, Hongchang and Pei, Jian and Huang, Heng},
  booktitle={Proceedings of the ACM Web Conference 2022},
  pages={1352--1361},
  year={2022}
}

@inproceedings{zhou2023hierarchical,
  title={Hierarchical knowledge graph learning enabled socioeconomic indicator prediction in location-based social network},
  author={Zhou, Zhilun and Liu, Yu and Ding, Jingtao and Jin, Depeng and Li, Yong},
  booktitle={Proceedings of the ACM Web Conference 2023},
  pages={122--132},
  year={2023}
}

@inproceedings{qu2023semi,
  title={Semi-decentralized federated ego graph learning for recommendation},
  author={Qu, Liang and Tang, Ningzhi and Zheng, Ruiqi and Nguyen, Quoc Viet Hung and Huang, Zi and Shi, Yuhui and Yin, Hongzhi},
  booktitle={Proceedings of the ACM Web Conference 2023},
  pages={339--348},
  year={2023}
}

@inproceedings{fan2019graph,
  title={Graph neural networks for social recommendation},
  author={Fan, Wenqi and Ma, Yao and Li, Qing and He, Yuan and Zhao, Eric and Tang, Jiliang and Yin, Dawei},
  booktitle={The world wide web conference},
  pages={417--426},
  year={2019}
}

@inproceedings{kanakia2019scalable,
  title={A scalable hybrid research paper recommender system for microsoft academic},
  author={Kanakia, Anshul and Shen, Zhihong and Eide, Darrin and Wang, Kuansan},
  booktitle={The world wide web conference},
  pages={2893--2899},
  year={2019}
}

@inproceedings{wang2022disencite,
  title={Disencite: Graph-based disentangled representation learning for context-specific citation generation},
  author={Wang, Yifan and Song, Yiping and Li, Shuai and Cheng, Chaoran and Ju, Wei and Zhang, Ming and Wang, Sheng},
  booktitle={Proceedings of the AAAI Conference on Artificial Intelligence},
  volume={36},
  number={10},
  pages={11449--11458},
  year={2022}
}

@article{wu2022nodeformer,
  title={Nodeformer: A scalable graph structure learning transformer for node classification},
  author={Wu, Qitian and Zhao, Wentao and Li, Zenan and Wipf, David P and Yan, Junchi},
  journal={Advances in Neural Information Processing Systems},
  volume={35},
  pages={27387--27401},
  year={2022}
}

@inproceedings{wang2020nodeaug,
  title={Nodeaug: Semi-supervised node classification with data augmentation},
  author={Wang, Yiwei and Wang, Wei and Liang, Yuxuan and Cai, Yujun and Liu, Juncheng and Hooi, Bryan},
  booktitle={Proceedings of the 26th ACM SIGKDD International Conference on Knowledge Discovery \& Data Mining},
  pages={207--217},
  year={2020}
}

@inproceedings{wei2023clnode,
  title={Clnode: Curriculum learning for node classification},
  author={Wei, Xiaowen and Gong, Xiuwen and Zhan, Yibing and Du, Bo and Luo, Yong and Hu, Wenbin},
  booktitle={Proceedings of the Sixteenth ACM International Conference on Web Search and Data Mining},
  pages={670--678},
  year={2023}
}

@inproceedings{zhaolearning,
  title={Learning on Large-scale Text-attributed Graphs via Variational Inference},
  author={Zhao, Jianan and Qu, Meng and Li, Chaozhuo and Yan, Hao and Liu, Qian and Li, Rui and Xie, Xing and Tang, Jian},
  booktitle={The Eleventh International Conference on Learning Representations}
}

@article{yan2023comprehensive,
  title={A comprehensive study on text-attributed graphs: Benchmarking and rethinking},
  author={Yan, Hao and Li, Chaozhuo and Long, Ruosong and Yan, Chao and Zhao, Jianan and Zhuang, Wenwen and Yin, Jun and Zhang, Peiyan and Han, Weihao and Sun, Hao and others},
  journal={Advances in Neural Information Processing Systems},
  volume={36},
  pages={17238--17264},
  year={2023}
}

@article{kingma2013auto,
  title={Auto-encoding variational bayes},
  author={Kingma, Diederik P},
  journal={arXiv preprint arXiv:1312.6114},
  year={2013}
}

@article{goodfellow2020generative,
  title={Generative adversarial networks},
  author={Goodfellow, Ian and Pouget-Abadie, Jean and Mirza, Mehdi and Xu, Bing and Warde-Farley, David and Ozair, Sherjil and Courville, Aaron and Bengio, Yoshua},
  journal={Communications of the ACM},
  volume={63},
  number={11},
  pages={139--144},
  year={2020},
  publisher={ACM New York, NY, USA}
}

@article{ho2020denoising,
  title={Denoising diffusion probabilistic models},
  author={Ho, Jonathan and Jain, Ajay and Abbeel, Pieter},
  journal={Advances in neural information processing systems},
  volume={33},
  pages={6840--6851},
  year={2020}
}

@article{mirza2014conditional,
  title={Conditional generative adversarial nets},
  author={Mirza, Mehdi},
  journal={arXiv preprint arXiv:1411.1784},
  year={2014}
}

@inproceedings{pumarola2020c,
  title={C-flow: Conditional generative flow models for images and 3d point clouds},
  author={Pumarola, Albert and Popov, Stefan and Moreno-Noguer, Francesc and Ferrari, Vittorio},
  booktitle={Proceedings of the IEEE/CVF Conference on Computer Vision and Pattern Recognition},
  pages={7949--7958},
  year={2020}
}

@article{cvae_sohn2015learning,
  title={Learning structured output representation using deep conditional generative models},
  author={Sohn, Kihyuk and Lee, Honglak and Yan, Xinchen},
  journal={Advances in neural information processing systems},
  volume={28},
  year={2015}
}

@inproceedings{schick2021s,
  title={It’s Not Just Size That Matters: Small Language Models Are Also Few-Shot Learners},
  author={Schick, Timo and Sch{\"u}tze, Hinrich},
  booktitle={Proceedings of the 2021 Conference of the North American Chapter of the Association for Computational Linguistics: Human Language Technologies},
  pages={2339--2352},
  year={2021}
}

@inproceedings{li2021prefix,
  title={Prefix-Tuning: Optimizing Continuous Prompts for Generation},
  author={Li, Xiang Lisa and Liang, Percy},
  booktitle={Proceedings of the 59th Annual Meeting of the Association for Computational Linguistics and the 11th International Joint Conference on Natural Language Processing (Volume 1: Long Papers)},
  pages={4582--4597},
  year={2021}
}

@inproceedings{lester2021power,
  title={The Power of Scale for Parameter-Efficient Prompt Tuning},
  author={Lester, Brian and Al-Rfou, Rami and Constant, Noah},
  booktitle={Proceedings of the 2021 Conference on Empirical Methods in Natural Language Processing},
  pages={3045--3059},
  year={2021}
}

@article{vaswani2017attention,
  title={Attention is all you need},
  author={Vaswani, A},
  journal={Advances in Neural Information Processing Systems},
  year={2017}
}

@article{cora_mccallum00automating,
  author = {McCallum, Andrew K. and Nigam, Kamal and Rennie, Jason and Seymore, Kristie},
  title = {Automating the Construction of Internet Portals with Machine Learning},
  journal = {Information Retrieval},
  volume = {3},
  number = {2},
  pages = {127--163},
  year = {2000},
  publisher = {Kluwer Academic Publishers},
  doi = {10.1023/A:1009953814988}
}

@inproceedings{ni2019justifying,
  title={Justifying recommendations using distantly-labeled reviews and fine-grained aspects},
  author={Ni, Jianmo and Li, Jiacheng and McAuley, Julian},
  booktitle={Proceedings of the 2019 Conference on Empirical Methods in Natural Language Processing and the 9th International Joint Conference on Natural Language Processing (EMNLP-IJCNLP)},
  pages={188--197},
  year={2019}
}

@inproceedings{finn2017model,
  title={Model-agnostic meta-learning for fast adaptation of deep networks},
  author={Finn, Chelsea and Abbeel, Pieter and Levine, Sergey},
  booktitle={International conference on machine learning},
  pages={1126--1135},
  year={2017},
  organization={PMLR}
}

@inproceedings{skorokhodovclass,
  title={Class Normalization for (Continual)? Generalized Zero-Shot Learning},
  author={Skorokhodov, Ivan and Elhoseiny, Mohamed},
  booktitle={International Conference on Learning Representations}
}

@article{xu2020attribute,
  title={Attribute prototype network for zero-shot learning},
  author={Xu, Wenjia and Xian, Yongqin and Wang, Jiuniu and Schiele, Bernt and Akata, Zeynep},
  journal={Advances in Neural Information Processing Systems},
  volume={33},
  pages={21969--21980},
  year={2020}
}

@inproceedings{cacheux2019modeling,
  title={Modeling inter and intra-class relations in the triplet loss for zero-shot learning},
  author={Cacheux, Yannick Le and Borgne, Herve Le and Crucianu, Michel},
  booktitle={Proceedings of the IEEE/CVF International Conference on Computer Vision},
  pages={10333--10342},
  year={2019}
}

@inproceedings{mishra2018generative,
  title={A generative model for zero shot learning using conditional variational autoencoders},
  author={Mishra, Ashish and Krishna Reddy, Shiva and Mittal, Anurag and Murthy, Hema A},
  booktitle={Proceedings of the IEEE/CVF Conference on Computer Vision and Pattern Recognition Workshops},
  pages={2188--2196},
  year={2018}
}

@inproceedings{schonfeld2019generalized,
  title={Generalized zero-and few-shot learning via aligned variational autoencoders},
  author={Schonfeld, Edgar and Ebrahimi, Sayna and Sinha, Samarth and Darrell, Trevor and Akata, Zeynep},
  booktitle={Proceedings of the IEEE/CVF Conference on Computer Vision and Pattern Recognition},
  pages={8247--8255},
  year={2019}
}

@article{elhoseiny2021cizsl++,
  title={Cizsl++: Creativity inspired generative zero-shot learning},
  author={Elhoseiny, Mohamed and Yi, Kai and Elfeki, Mohamed},
  journal={arXiv preprint arXiv:2101.00173},
  year={2021}
}

@inproceedings{shen2020invertible,
  title={Invertible zero-shot recognition flows},
  author={Shen, Yuming and Qin, Jie and Huang, Lei and Liu, Li and Zhu, Fan and Shao, Ling},
  booktitle={Computer Vision--ECCV 2020: 16th European Conference Proceedings, Part XVI 16},
  pages={614--631},
  year={2020}
}

@inproceedings{reimers-2019-sentence-bert,
  title = "Sentence-BERT: Sentence Embeddings using Siamese BERT-Networks",
  author = "Reimers, Nils and Gurevych, Iryna",
  booktitle = "Proceedings of the 2019 Conference on Empirical Methods in Natural Language Processing",
  month = "11",
  year = "2019",
  publisher = "Association for Computational Linguistics",
  url = "https://arxiv.org/abs/1908.10084",
}









\end{document}